\documentclass[lettersize,journal]{IEEEtran}
\usepackage[table]{xcolor}
\usepackage[utf8]{inputenc}
\usepackage{amsmath,amsfonts}
\usepackage{algorithmic}
\usepackage{algorithm}
\usepackage{array}
\usepackage{subfigure}
\usepackage{textcomp}
\usepackage{stfloats}
\usepackage{url}
\usepackage{verbatim}
\usepackage{graphicx}
\usepackage{cite}
\usepackage{color}
\usepackage{booktabs}
\usepackage{tabularx}
\usepackage{multirow}
\usepackage{utfsym}
\usepackage{hyperref}
\usepackage{makecell}

\definecolor{barrier}{RGB}{242,122,52}
\definecolor{bicycle}{RGB}{237,186,191}
\definecolor{bus}{RGB}{255,255,79}
\definecolor{car}{RGB}{0,134,207}
\definecolor{constveh}{RGB}{0,222,222}
\definecolor{motorcycle}{RGB}{161,141,0}
\definecolor{pedestrian}{RGB}{230,32,20}
\definecolor{trafficcone}{RGB}{218,207,125}
\definecolor{trailer}{RGB}{109,58,0}
\definecolor{truck}{RGB}{160,89,255}
\definecolor{drivesurf}{RGB}{255,0,255}
\definecolor{otherflat}{RGB}{122,122,122}
\definecolor{sidewalk}{RGB}{46,10,71}
\definecolor{terrain}{RGB}{158,239,112}
\definecolor{manmade}{RGB}{201,201,201}
\definecolor{vegetation}{RGB}{0,158,0}

\newcommand{\sq}[1]{\textcolor{#1}{\rule{0.9em}{0.9em}}}

\newcommand{\RotCol}[2]{%
  \makecell[b]{     
    \rotatebox{90}{\footnotesize #1}\\[2pt]
    \sq{#2}%
  }%
}

\begin{document}

\title{HV-BEV: Decoupling Horizontal and Vertical Feature Sampling for Multi-View 3D Object Detection}

\author{Di Wu, Feng Yang, \textit{Member, IEEE,} Benlian Xu, Pan Liao, Wenhui Zhao, Dingwen Zhang, \textit{Member, IEEE}

\thanks{This work was supported in part by the National Natural Science
Foundation of China (No. 61374159, U24A20263), Shaanxi Natural Fund (No. 2018MJ6048), Space Science and Technology Fund, the Foundation of CETC Key Laboratory of Data Link Technology (CLDL-20182316, CLDL20182203), and the Suzhou municipal science and technology plan project (No. SYG202351). (Corresponding author: Feng Yang and Benlian Xu.)}

\thanks{D. Wu, Feng Yang, Pan Liao and Wenhui Zhao are with the school of automation, Northwestern Polytechnical University, Xi'an Shanxi 710072, China (e-mail: wu\_di821@mail.nwpu.edu.cn; yangfeng@nwpu.edu.cn; liaopan@mail.nwpu.edu.cn; zwh2024202513@mail.nwpu.edu.cn; zdw2006yyy@nwpu.edu.cn).}
\thanks{Benlian Xu is with the school of electronic and information engineering, Suzhou University of Science and Technology, Suzhou Jiangsu 215009, China (e-mail:xu\_benlian@usts.edu.cn).}
        }

\markboth{Manuscript}%
{Shell \MakeLowercase{\textit{et al.}}: A Sample Article Using IEEEtran.cls for IEEE Journals}

\maketitle

\begin{abstract}
The application of vision-based multi-view environmental perception system has been increasingly recognized in autonomous driving technology, especially the BEV-based models. Current state-of-the-art solutions primarily encode image features from each camera view into the BEV space through explicit or implicit depth prediction. However, these methods often overlook the structured correlations among different parts of objects in 3D space and the fact that different categories of objects often occupy distinct local height ranges. For example, trucks appear at higher elevations, whereas traffic cones are near the ground. In this work, we propose a novel approach that decouples feature sampling in the \textbf{BEV} grid queries paradigm into \textbf{H}orizontal feature aggregation and \textbf{V}ertical adaptive height-aware reference point sampling (HV-BEV), aiming to improve both the aggregation of objects' complete information and awareness of diverse objects' height distribution. Specifically, a set of relevant neighboring points is dynamically constructed for each 3D reference point on the ground-aligned horizontal plane, enhancing the association of the same instance across different BEV grids, especially when the instance spans multiple image views around the vehicle. Additionally, instead of relying on uniform sampling within a fixed height range, we introduce a height-aware module that incorporates historical information, enabling the reference points to adaptively focus on the varying heights at which objects appear in different scenes. Extensive experiments validate the effectiveness of our proposed method, demonstrating its superior performance over the baseline across the nuScenes dataset. Moreover, our best-performing model achieves a remarkable 50.5\% mAP and 59.8\% NDS on the nuScenes testing set. The code is available at \url{https://github.com/Uddd821/HV-BEV}.
\end{abstract}


\section{Introduction}

\begin{figure}[!t]
\centering
\includegraphics[width=3.5in]{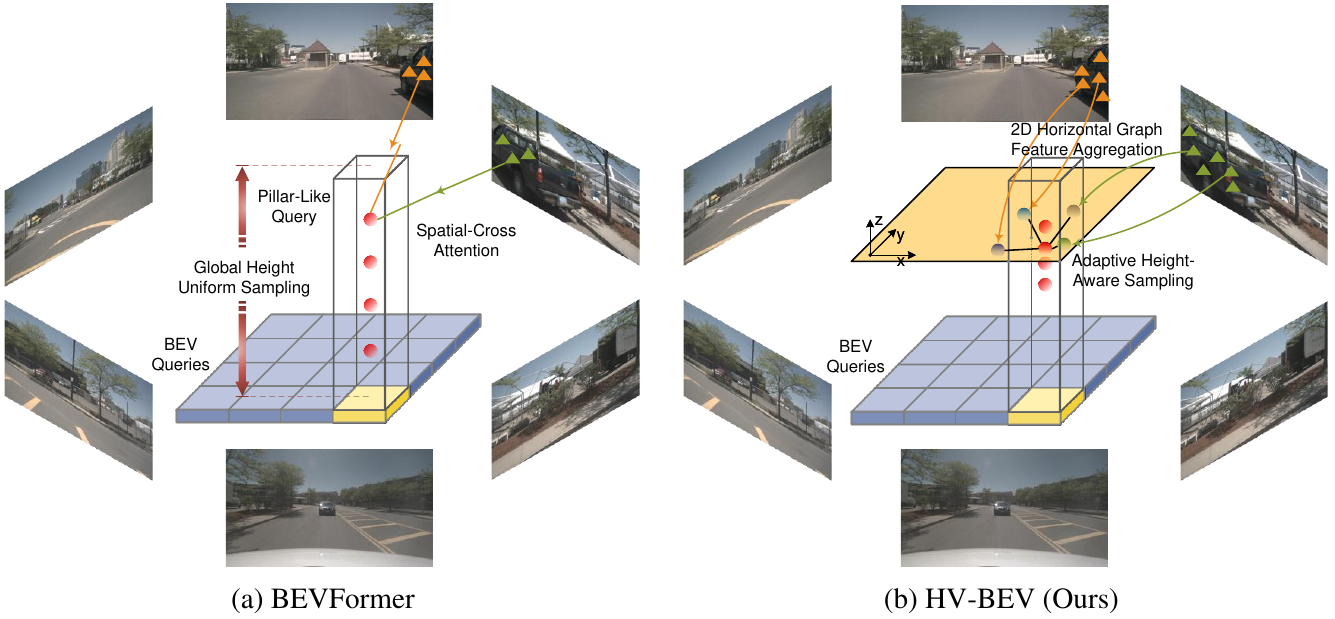}
\caption{The multi-view feature sampling comparison between (a) BEVFormer and (b) our HV-BEV. BEVFormer extracts local area features by uniformly sampling several reference points across the global height range within each pillar-like query and projecting them onto corresponding views. In contrast, HV-BEV adopts adaptive height-aware sampling and constructs a graph for each reference point on the horizontal plane, aggregating features from relevant regions to enhance feature coherence.}
\label{fig_1}
\end{figure}

\IEEEPARstart{3}{D} object detection is a fundamental task in intelligent vehicle perception systems, focused on estimating the locations, sizes, and categories of key objects in traffic scenarios—such as vehicles, pedestrians, cyclists, and traffic cones—in the ego-vehicle coordinate system \cite{mao20233d}. Although LiDAR provides direct distance measurements that facilitate 3D detection \cite{zhou2018voxelnet, liu2024multi, li2024scnet3d}, its high cost and limited maintainability make cameras—cost-effective devices capable of capturing rich semantic information—an essential choice for deployment. By strategically positioning cameras at various angles around the vehicle, the system can achieve comprehensive perception of its surroundings. Consequently, multi-view visual 3D object detection remains one of the central focuses of ongoing research \cite{ma2024vision, zhang2023occformer}. 

Recently, BEV(bird's eye view)-based 3D perception has gained widespread attention \cite{xie2022m, li2024fast, huang2024gaussianformer} due to the ability of BEV feature representation to model a complete, scale-consistent environment around the vehicle within a unified coordinate system, making it more suitable for downstream tasks such as path planning and control. BEV-based 3D object detection can be broadly categorized into two major methods: explicit and implicit depth estimation \cite{ma2024vision}, based on how 2D features are transformed into 3D space. Explicit depth-based methods primarily utilize depth estimation networks to predict depth distribution in perspective views (PV), transforming PV feature maps into 3D space using determined camera parameters \cite{huang2021bevdet, huang2022bevdet4d, li2023bevdepth, li2023bevstereo}. Instead of the intermediate depth estimation step, implicit methods rely on sophisticated view transformation networks to extract 3D information embedded in the 2D images without the need for an depth map \cite{wang2022detr3d, li2022bevformer, yang2023bevformer, liu2022petr, liu2023petrv2}. 

Most of these existing methods focus on minimizing the feature localization errors caused by depth estimation inaccuracies, while neglecting the 3D structural information of real-world objects. In other words, they pay little attention to the global representation of regions associated with an object when encoding BEV features, resulting in insufficient feature aggregation. While some works have addressed this issue with promising results \cite{chen2022graph, chen2024graph, lin2022sparse4d}, they are predominantly centered around sparse object query-based schemes, which are specifically designed for the single task of 3D object detection and are difficult to extend to other tasks, i.e., map segmentation, by simply replacing the task head or transform into multi-task models. In addition, most BEV-based methods either "splat" frustum features through Voxel/Pillar Pooling \cite{philion2020lift, huang2021bevdet, li2023bevdepth} or directly cross-encode image features into a flattened BEV grid plane \cite{li2022bevformer}, leading to the collapse of the height dimension and causing the model to overlook crucial height information. Different object categories tend to occupy distinct local height ranges \cite{chi2023bev}, yet treating all vertical positions equally during feature sampling can lead to ineffective aggregation. For instance, traffic cones typically reside near the ground, but sampling features uniformly along the entire height axis within their corresponding pillar introduces a large amount of irrelevant information, resulting in wasted computation and degraded detection performance. To address this, the network should adaptively focus on the local height regions where objects are likely to appear and, in conjunction with horizontally aggregated features, construct more compact and discriminative 3D representations.

In this paper, we take BEVFormer \cite{li2022bevformer}, a representative work of the BEV grid query-based approach, as our baseline and propose a comprehensive 3D object detection framework named HV-BEV that decouples multi-view feature sampling in the horizontal BEV plane and vertical height direction. In the vanilla BEVFormer, multi-view feature sampling is achieved by projecting the 3D reference points located at the center of the corresponding cell region of each BEV query and uniformly sampled across the global height range to the hit views and applying deformable attention \cite{zhu2021deformable}, as shown in Fig. 1(a). This sampling approach retrieves the corresponding image features for each BEV grid region from different views and implicitly modeling height information; However, it does not fully account for the structured features of 3D objects, particularly the cohesive nature of height information and the interdependencies across different dimensions. We identified that large or near-ego objects are sometimes missed or inaccurately estimated due to occupying multiple BEV grid regions or appearing in multiple camera views through observation. This phenomenon is primarily caused by insufficient feature aggregation in the horizontal dimension. To address this, we propose a dynamic horizontal cross-view feature aggregation (DHCA) module, which constructs a set of neighboring points for each 3D reference point on the horizontal BEV plane to aggregate global features of the object from different grid regions and views, seen in Fig. 1(b). However, relying solely on horizontal feature aggregation is not enough, as different objects do not exist along a single horizontal line; rather, their heights vary with scene context and object class. Motivated by this, we introduce an adaptive height-aware sampling strategy as shown in Fig. 1(b). The height distributions from aligned historical BEV features and current BEV queries are fused and guide vertical height sampling of 3D reference points, allowing the network to adaptively focus on the most relevant height for each object. The training of this module is done in a sparse way, where discrete height distributions are generated based on the centers of 3D annotated bounding boxes to supervise object height estimation. By decoupling feature sampling in different dimensions, the model can concentrate on information of value across various directions, yielding a more efficient global feature sampling strategy. Notably, while this paper evaluates the performance of HV-BEV using 3D object detection as the primary task for model training and prediction, our approach can be easily extended to other BEV-based perception tasks, such as map segmentation and lane detection. 

Our contributions can be concluded as follows:
\begin{itemize}
\item We propose a multi-view 3D object detection method based on the BEV grid query paradigm, which decouples horizontal and vertical feature sampling to fully exploit 3D spatial information, achieving superior performance. \item We design a dynamic horizontal cross-view feature aggregation module that fuses relevant structured features of the object from different camera images and grid cell regions and an adaptive height-aware sampling strategy in the vertical dimension to guide the network in identifying relevant object heights of interest. \item Extensive experiments conducted on the nuScenes and Lyft datasets to validate the effectiveness of the proposed method. Our HV-BEV outperforms the baseline model across all evaluated metrics and exhibits outstanding performance within the camera-based 3D object detection community. 
\end{itemize}

\section{Related Work}
\subsection{Explicit 3D Object Detection}
The explicit-based methods utilize camera imaging principles to directly estimate the depth of each pixel in the image, enabling precise construction of 3D features. LSS \cite{philion2020lift} is a foundational work in this category, predicting the discrete depth distribution \cite{reading2021categorical} of pixel points and, through a novel outer product with 2D image features, lifting these features into a frustum-like 3D perspective space to simulate feature point distribution along the line of sight. The BEVDet series \cite{huang2021bevdet, huang2022bevdet4d} follows the LSS paradigm, introducing a versatile framework for 3D object detection composed of four main modules: an image encoder, a view transformation module, a BEV encoder, and a task-specific head. BEVDepth \cite{li2023bevdepth} introduces lidar point cloud supervision for the depth estimation and encodes camera parameters into the network, enhancing robustness and accuracy in depth prediction and thereby improving detection performance. BEVStereo \cite{li2023bevstereo} enhances depth estimation by utilizing two-frame stereo images, predicting depth from both single-frame features and temporal stereo information. SOLOFusion \cite{park2022time} further enhances single-frame depth prediction by generating a cost volume from long-term historical images. FB-BEV \cite{li2023fb} introduces a depth-aware back-projection process to refine the sparse BEV representation generated by the LSS-style forward projection. BEVNeXt \cite{li2024bevnext} argues that long-term fusion strategies, like those in SOLOFusion \cite{park2022time, han2024exploring}, are constrained by limited receptive fields; thus, it expands the receptive field and enhances object-level BEV features derived through backward projection by incorporating CRF-modulated depth embeddings. Since these methods rely on depth prediction to accomplish the 2D-to-3D transformation, most approaches aim to enhance BEV representation by improving depth prediction accuracy through various strategies. However, depth estimation inevitably introduces errors that accumulate as network layers deepen. 

\subsection{Implicit 3D Object Detection}
Following the successful application of transformer \cite{vaswani2017attention, brown2020language, liu2021swin} in computer vision \cite{dosovitskiy2020image, carion2020end, zhao2024detrs}, many query-based 3D detection frameworks have emerged, leveraging the transformer’s powerful feature encoding capabilities to eliminate the need for explicit depth prediction step. These frameworks can be categorized into sparse query-based methods, like DETR3D \cite{wang2022detr3d}, which extract relevant features from 2D images using 3D queries, and dense query-based methods, like BEVFormer \cite{li2022bevformer, yang2023bevformer}, which predefine BEV grid queries to query corresponding BEV features from multi-view images. Considering the limitations of DETR3D's single-point sampling in capturing global representations, the PETR series \cite{liu2022petr, liu2023petrv2} introduces a streamlined framework that bypasses direct 2D-to-3D transformation. Instead, it encodes 2D features with 3D positional embeddings, enabling the network to capture 3D spatial awareness. Graph-DETR3D \cite{chen2022graph} identifies "truncated instances" at image boundary regions as a primary bottleneck limiting the detection performance of DETR3D \cite{wang2022detr3d}. To mitigate this issue, it introduces a dynamic 3D graph structure between each object queries and 2D feature maps, further extending this spatial graph into a 4D spatio-temporal space \cite{chen2024graph}. The Sparse4D series \cite{lin2022sparse4d, lin2023sparse4d} generates a set of keypoints for each instance’s anchor box, leveraging these keypoints to aggregate instance features across multiple time steps, scales, and viewpoints. While sparse query-based methods offer a balance between accuracy and speed, dense query-based approaches are better suited for multi-task extension due to their unified BEV representation. This paper aims to enhance the global spatial awareness of such models by using BEVFormer as the baseline.

\subsection{Height-Aware 3D Object Detection}
Height information is often overlooked in BEV-based 3D object detection, yet it contains valuable object attribute features; for example, traffic cones are typically close to the ground, while buses are relatively tall. BEV-SAN \cite{chi2023bev} is the first to incorporate height-specific features of different objects within the network, using a height histogram of LiDAR points to define the upper and lower bounds of local BEV slices. This approach captures features across various heights in BEV space and fuses them with global BEV features, proving effective in increasing detection accuracy. HeightFormer \cite{wu2024heightformer} further demonstrates the equivalence of predicting height in BEV and predicting depth in images for 2D-to-3D mapping, presenting a BEV detection framework that explicitly models height as an alternative to depth-based methods. BEVHeight \cite{yang2023bevheight, yang2023bevheight++} focuses on improving 3D object detection performance for roadside applications, recognizing that depth-based methods, when directly transferred, often perform poorly with roadside cameras. This limitation arises because the depth of distant objects and the ground depth are nearly indistinguishable from a roadside camera’s perspective. To address this, BEVHeight leverages the more stable information provided by object height, enabling robust detection outcomes. OC-BEV \cite{qi2024ocbev} refines the reference point sampling in BEVFormer by introducing an object focused sampling mechanism. While it increases the sampling density in high-frequency height regions by presetting a local height range and does not fully exploit object height embeddings within the network. This paper explicitly extracts height information from BEV features to adaptively determine the heights of interest, facilitating a feature sampling method that is more generalizable and adaptable to diverse scenes.

\section{Methods}

\subsection{Revisiting to BEVFormer-Style Feature Sampling}
To clearly present our proposed framework, we first review the feature sampling method in the baseline BEVFormer. It initially employs a backbone network to extract multi-view feature maps $F_t=\{F_t^i\}_{i=1}^{N_{\text{view}}}$ from multi-view images, where $F_t^i$ is the features of the $i$-th view at current timestamp $t$. It then introduces a BEV feature encoder structured similarly to a transformer decoder, consisting of multiple encoder layers in sequence. Each layer takes grid-shaped BEV queries $Q\in\mathbb{R}^{H\times W\times C}$ as input, where $H, W$ denotes the shape of BEV plane and $C$ is the channel number, and progressively refines the current BEV features $B_t$ through temporal self-attention with historical BEV features $B_{t-1}$, spatial cross-attention with the multi-view feature maps $F_t$, and additional transformer decoder components. Specifically, each query $Q_p\in\mathbb{R}^{1\times C}$ of $Q$, representing a rectangular region centered at $p\in(x,y)$ in the BEV plane, first leverages temporal self-attention (TSA) to model temporal relationships from aligned historical BEV features $B'_{t-1}$ to capture the sequential associations of objects over time:

\begin{equation}
\begin{array}{l}
\text{TSA}(Q_p,\{Q,B'_{t-1}\})=\sum\limits_{V\in\{Q,B'_{t-1}\}}\text{DefAttn}(Q_p,p,V)
\end{array}
\tag{1},
\end{equation}
where DefAttn$(\cdot)$ is the deformable attention \cite{zhu2021deformable} and $\{\cdot\}$ denotes the concatenation operation of two tensors. After applying a residual connection and a layer normalization (Add\&Norm) \cite{vaswani2017attention} to the output BEV queries, spatial cross-attention (SCA) is then used to aggregate spatial information from multi-view features:

\begin{equation}
\begin{array}{l}
\text{SCA}(Q_p,F_t)=\dfrac{1}{\vert \mathcal{V}_{hit}\vert}\sum\limits_{i\in\mathcal{V}_{\text{hit}}}\sum\limits_{j=1}^{N_{\text{ref}}}\text{DefAttn}(Q_p,\mathcal{P}(p,i,j),F_t^i)
\end{array}
\tag{2},
\end{equation}
where $N_{\text{ref}}$ denotes the number of 3D reference points uniformly sampled within a predetermined height range for each query $Q_p$, centered at the horizontal coordinate $p(x,y)$. $\mathcal{P}(p, i, j)$ represents the projection of the $j$-th reference point in $Q_p$ onto the 2D image plane of the $i$-th camera view using the camera intrinsic matrix, and $\mathcal{V}_{hit}$ denotes the set of views in which the reference point is successfully projected. Following the temporal and spatial feature interactions via TSA and SCA, the resulting BEV queries are subjected to passed through an Add\&Norm layer, then a forward propagation network (FFN) \cite{vaswani2017attention} and the last Add\&Norm layer, yielding the final refined BEV features $B_t$.

\subsection{Overall Framework}
\begin{figure*}[!t]
\centering
\includegraphics[width=7in]{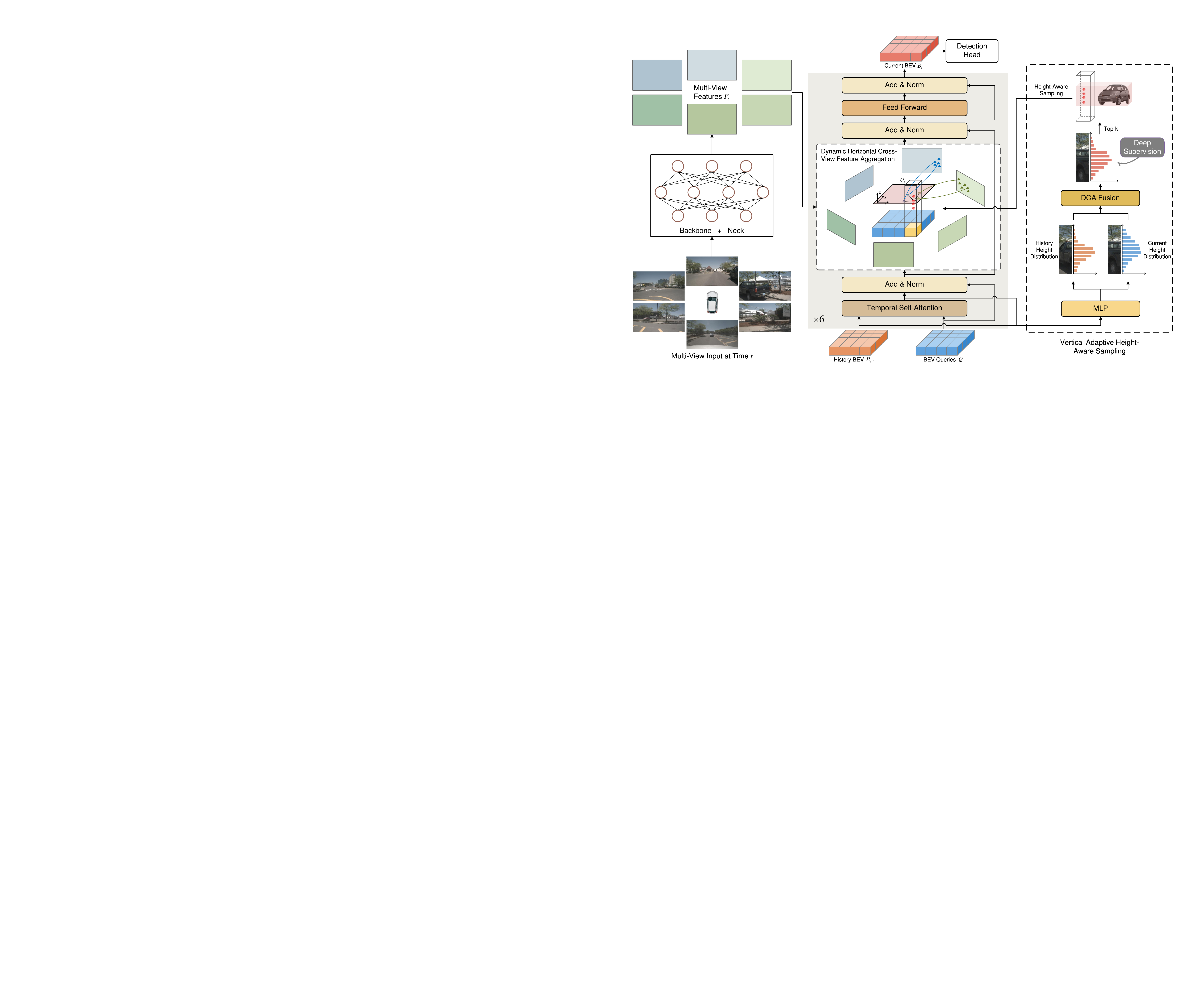}
\caption{Overall architecture of HV-BEV. Aligning with BEVFormer, we begin by using an image encoder to extract multi-view image features. Within the six-layer BEV encoder, we introduce two decoupled modules that respectively handle intrinsic height awareness and cross-view feature aggregation. The \textit{Vertical Adaptive Height-Aware module} is designed to capture the potential height distribution of objects within each BEV grid, aiding in the sampling of 3D reference points. Leveraging these height-aware 3D reference points, the \textit{Dynamic Horizontal Cross-View Feature Aggregation module} constructs a set of neighboring points on the BEV plane for each one to explore spatial features from neighboring regions.}
\label{fig_2}
\end{figure*}

As illustrated in Fig. 2, our HV-BEV follows the 6-layer encoder architecture of BEVFormer, with the primary difference being an improved multi-view feature sampling method within each layer. Specifically, we first employ an image encoder—comprising a backbone, such as ResNet \cite{he2016deep}, and a neck, like FPN \cite{lin2017feature}—to extract multi-view features $F_t$ from the multi-view images at time $t$, which then serve as input for the dynamic horizontal cross-view feature aggregation module in the next BEV encoder. 

A BEV encoder layer receives learnable BEV queries $Q$ and the preserved BEV features $B_{t-1}$ from the previous time step as input. We first apply a standard temporal self-attention layer to query temporal information from historical BEV as Eq.(1). Following this, a vertical height-aware sampling (VHA) module (Sec. \uppercase\expandafter{\romannumeral3}-C) integrates the predicted height distribution from each $Q_p$ of the current BEV queries, obtained through a shared MLP layer, with the historical height distribution generated from $B_{t-1}$. From this fused distribution, the module samples $N_{ref}$ discrete height values using a top-k operation and assigns them as the height coordinates of the $N_{ref}$ 3D reference points corresponding to each query $Q_p$. Notably, by introducing deep supervision into this module, the BEV queries are guided to focus on the height ranges where objects are most likely to appear. In the dynamic horizontal cross-view feature aggregation (DHCA) module (Sec. \uppercase\expandafter{\romannumeral3}-D), each 3D reference point learns a set of neighboring points determined by learnable 2D offsets within its height-specific BEV plane. All points in this graph are then projected onto multi-view images to fully aggregate spatial features from the hit views. This approach allows for comprehensive integration of neighboring regions' spatial information that may appear in different views, further refining BEV features. After passing through additional standard layers (FFN, Add\&Norm), the refined BEV features are fed as input queries to the successive BEV encoder layer and iteratively enhanced, producing the final BEV features, which are then utilized by task-specific heads for 3D detection and other perception tasks.

\subsection{Vertical Adaptive Height-Aware Sampling}
In HeightFormer\cite{wu2024heightformer}, the authors use attention weights to compute a weighted average of predefined anchor heights within a BEV sample, thereby visualizing how common SCA implicitly encodes height information. Instead of sampling at fixed heights across the entire range, we propose an adaptive height-aware sampling method that enables the model to dynamically adjust the optimal sampling heights based on the specific scene characteristics.

Through TSA, the current BEV queries $Q$ can retrieve temporal information from the historical BEV $B_{t-1}$ to generate the fused queries $Q'\in\mathbb{R}^{H\times W\times C}$. To further integrate historical height information for predicting the current height distribution, we first employ a shared MLP layer to independently predict the discrete historical and current height distributions from $B_{t-1}$ and $Q'$, resulting in $H^{hist}, H^{cur} \in \mathbb{R}^{H \times W \times D}$, where $D$ denotes the number of discrete height bins over the global height range. However, due to ego-motion, it is necessary to identify the height distribution at the previous position $p_{t-1}$ corresponding to the grid region centered at the current location $p$. This is achieved by establishing a mapping $\mathcal{M}$ from time $t$ to time $t-1$, defined as follows:

\begin{equation}
\begin{array}{l}
\mathcal{M}: p_{t-1}\leftarrow f_{t,t-1}(p,\mathbf{R}_{t,t-1},\mathbf{T}_{t,t-1})
\end{array}
\tag{3},
\end{equation}

\begin{equation}
\begin{array}{l}
f_{t,t-1}(p,\mathbf{R}_{t,t-1},\mathbf{T}_{t,t-1})=\mathbf{R}_{t,t-1}\cdot p+\mathbf{T}_{t,t-1}
\end{array}
\tag{4},
\end{equation}
where $\mathbf{R}_{t,t-1} \in \mathbb{R}^{2 \times 2}$ and $\mathbf{T}_{t,t-1} \in \mathbb{R}^{2 \times 1}$ denote the rotation matrix around the vertical $z$ axis and the horizontal translation vector from time $t$ to $t{-}1$, respectively. Through this mapping, the historical height distribution at $p_{t-1}$ can be warped into the coordinate system at time $t$, resulting in the warped historical height distribution $H^{hist}_{t-1\rightarrow t} \in \mathbb{R}^{H \times W \times D}$. Subsequently, the historical and current height information is further fused.

\begin{figure}[!t]
\centering
\includegraphics[width=3.5in]{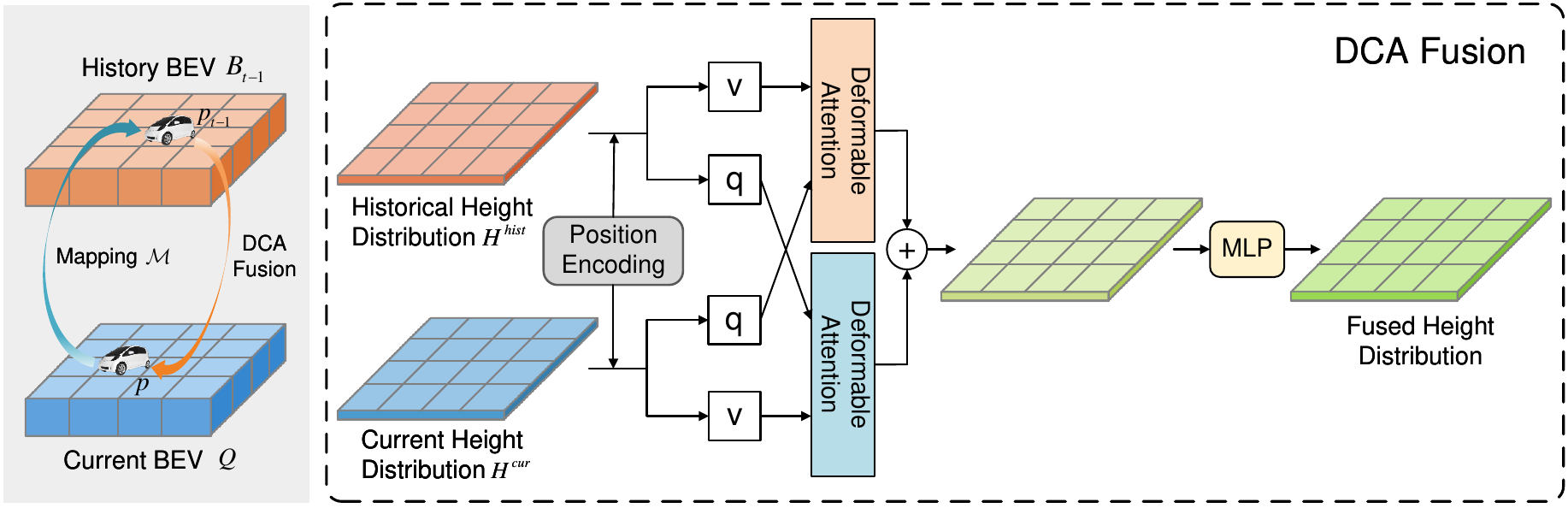}
\caption{The dual-branch cross-attention (DCA) fusion strategy between the current height distribution and historical height distribution. "q" represents the query, and "v" represents the value. In this figure, we omit the key because it corresponds to the center position $p$ of the grid in the counterpart BEV space that is associated with the query, as determined by $\mathcal{M}$.}
\label{fig_3}
\end{figure}

Unlike depth, the height of a moving object remains constant, allowing us to refine the current height distribution prediction by merging $H^{cur}$ and $H^{hist}_{t-1\rightarrow t}$. Inspired by \cite{chi2023bev}, we introduce a dual-branch cross-attention (DCA) fusion strategy to effectively integrate these height distributions. As shown in Fig.3, it consists of two deformable attention branches. In each branch, the height distribution $H_p^{cur/hist}$ at each grid cell serves as the query, aggregating information related to the grid center position $p$ from $H^{hist/cur}$ in another branch (For simplicity, the subscript of $H^{hist}_{t-1\rightarrow t}$ is omitted here.):

\begin{equation}
\begin{array}{l}
\text{DCA}(H_p^{cur/hist},H^{hist/cur})=\\ \hspace{1.2cm}\text{DefAttn}(H_p^{cur/hist},p,H^{hist/cur})
\end{array}
\tag{5}.
\end{equation}

The output distributions from both branches are then summed and passed through an MLP layer to generate the fused height distribution $H^{fus}\in\mathbb{R}^{H\times W\times D}$. For each resulting height distribution $H_p^{fus}$, we apply a top-k function to sample $N_{ref}$ discrete height values, which are subsequently used as the height coordinates of the 3D reference points in the dynamic horizontal cross-view feature aggregation. 

During training, we apply sparse supervision to $H^{fus}$ through utilizing the height of the annotated 3D bounding box center. Specifically, to facilitate the calculation of the discrepancy between the ground-truth height and the discretized predicted height distribution, we first derive the discretized ground-truth height distribution $H_p^{gt}$ corresponding to each BEV grid region. Since an object’s height is inherently a range rather than a fixed value, we employ a Gaussian kernel $K$ to compute the differences between the annotated center height of an object and the predefined height bins. Height bins closer to the center are assigned higher probabilities. For convenience, we use an indicator $I_p$ to denote whether the cell grid centered at $p$ falls within the object's region. The probability corresponding to the $m$-th height bin of $H_p^{gt}$ is calculated as:

\begin{equation}
\begin{array}{l}
H_{p,m}^{gt}=\begin{cases}
\dfrac{K(z_{gt},z_m)}{\sum\limits_{n=1}^D K(z_{gt},z_n)}, &I_p=1\\
1/D, &I_p=0
\end{cases}
\end{array}
\tag{6},
\end{equation}
\begin{equation}
\begin{array}{l}
K(z_{gt},z_m)=e^{-\dfrac{\vert z_{gt}-z_m\vert^2}{2\sigma^2}}
\end{array}
\tag{7},
\end{equation}
where $z_{gt}$ denotes the height of ground truth center, $z_m$ represents the center value of the $m$-th height bin, and $\sigma$ is the standard deviation controlling the spread of the Gaussian kernel. When $I_p = 0$, assigning equal probability to each height bin encourages the network to attend uniformly to the entire height range, allowing it to handle potentially missed or unannotated objects. The height distribution is supervised as follows: 

\begin{equation}
\begin{array}{l}
L_{hgt}=-\dfrac{1}{HW}\sum\limits_{p\in \mathcal{P}_{\text{bev}}}H_{p}^{gt} \log H_{p}^{fus}
\end{array}
\tag{8}.
\end{equation}
Here, the grid set $\mathcal{P}_{\text{bev}}=\{(i,j)\vert i=1,\dots,H;j=1,\dots,W\}$ is defined as the collection of all grid center positions on the BEV plane.

\subsection{Dynamic Horizontal Cross-View Feature Aggregation}
The standard spatial cross-attention (SCA) mechanism projects each 3D reference point to all image coordinate systems, summing features extracted from all hit views; this approach partially aggregates cross-view spatial information for 3D points that can be projected into adjacent views after coordinate transformation. However, for reference points projected into only one view, the imaging information of their associated objects may span multiple views, leading to inadequate feature aggregation. In this section, we present a dynamic horizontal cross-view neighborhood aggregation (DHCA) module, which learns a set of relevant neighboring points on the horizontal plane for each 3D reference point, forming a graph-like structure centered around the reference point. For clarity of presentation, we utilize the adjacency matrix and edges in a graph structure to represent the relationships between each reference point and its neighboring points. As illustrated in Fig.4, this module refines each point’s feature representation by integrating sampled features from neighboring points, thereby enhancing spatial correlations and improving global feature aggregation.

Formally, for any 3D reference point $p_{3d}\in(x,y,z)$ in $Q_p$, where $(x,y)$ represents the horizontal center coordinates of the corresponding grid region and $z$ is the vertical height coordinate, we define a set of neighboring points $\mathbf{V}=\{v_k\}_{k=1}^M$ centered on $p_{3d}$, where $M$ denotes the number of $p_{3d}$'s neighbors. And $\mathbf{X}\in\mathbb{R}^{M\times C}$ represents the feature embeddings of all neighboring points, where the $k$-th feature vector $\mathbf{x}_k\in\mathbb{R}^C$ corresponds to the attribute of $v_k$. Furthermore, each neighboring point $v_k(k=1,\dots,M)$ is learned from a linear layer:

\begin{equation}
\begin{array}{l}
v_k=p_{3d}+\bigtriangleup p_{xy}^k=(x+\bigtriangleup x_k,y+\bigtriangleup y_k,z)
\end{array}
\tag{9},
\end{equation}
\begin{equation}
\begin{array}{l}
\bigtriangleup p_{xy}^k=\text{Linear}(Q_p)\in \mathbb{R}^2
\end{array}
\tag{10},
\end{equation}
where $\bigtriangleup p_{xy}^k=(\bigtriangleup x_k,\bigtriangleup y_k)$ represent the sampling offsets along two dimensions of the horizontal plane. Notably, since the distribution of associated features may vary across different heights of an object, the sampling offsets for different reference points in $Q_p$ are obtained individually through the shared linear layer. The feature vector of $p_{3d}$, denoted as $\mathbf{x}_p$, is obtained via deformable attention, similar to the vanilla SCA:

\begin{equation}
\begin{array}{l}
\mathbf{x}_p=\dfrac{1}{\vert \mathcal{V}_{hit}\vert}\sum\limits_{i\in\mathcal{V}_{\text{hit}}}\text{DefAttn}(Q_p,\mathcal{P}(p_{3d},i),F_t^i)
\end{array}
\tag{11},
\end{equation}
where $\mathcal{P}(p_{3d},i)$ is the 2D point on $i$-th view projected from $p_{3d}$ via the known instrinsic parameter matrix $\mathbf{I}_i\in\mathbb{R}^{3\times4}$ of the $i$-th camera, which can be written as: 

\begin{equation}
\begin{array}{l}
\mathcal{P}(p_{3d},i)=(x_i,y_i)
\end{array}
\tag{12},
\end{equation}
\begin{equation}
\begin{array}{l}
z_i\cdot[x_i\text{ }y_i\text{ }1]^T=\mathbf{I}_i\cdot[x\text{ }y\text{ }z\text{ }1]^T
\end{array}
\tag{13},
\end{equation}
where $(x_i,y_i,z_i)$ denotes the pixel $(x_i,y_i)$ with depth $z_i$ projected onto the $i$-th camera image.

To optimize computational efficiency, the features of the neighboring points $v_k(k=1, \dots, M)$ are extracted from the corresponding image features using bilinear interpolation: 

\begin{equation}
\begin{array}{l}
\mathbf{x}_k=\dfrac{1}{\vert \mathcal{V}_{hit}\vert}\sum\limits_{i\in\mathcal{V}_{\text{hit}}}\text{Bilinear}(\mathcal{P}(v_k,i),F_t^i)
\end{array}
\tag{14}.
\end{equation}
$\mathcal{P}(v_k,i)$ represents the projection of $v_k$ onto the $i$-th camera view using the camera intrinsic matrix, and $\mathcal{V}_{hit}$ denotes the hit views.

After obtaining the attributes of all neighboring points, we update the centered reference point feature via:
\begin{equation}
\begin{array}{l}
\mathbf{x}'_p=\mathbf{x}_p+\sum\limits_{k=1}^M\mathbf{w}_{k}\cdot\mathbf{x}_k
\end{array}
\tag{15}.
\end{equation}

Here, $\mathbf{w}_k$ represents the edge weight vector between $p_{3d}$ and its neighbors, learned through a linear layer followed by a softmax function. Finally, the features of all $N_{ref}$ reference points are aggregated via a sum operation, yielding the refined query $Q_p$ as the output of DHCA module, fully enriched with spatial information.

\begin{figure}[!t]
\centering
\includegraphics[width=3.5in]{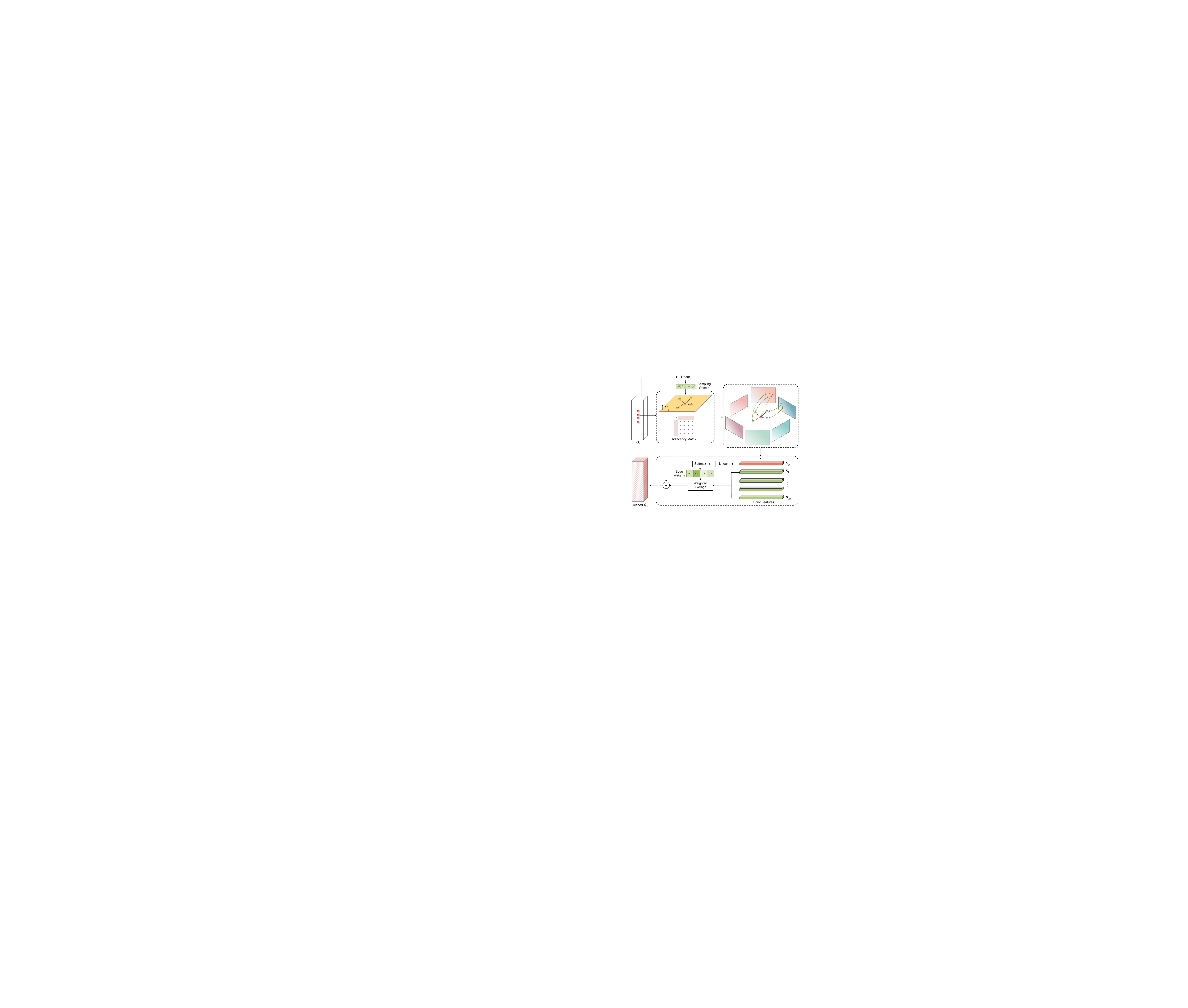}
\caption{The process of the dynamic horizontal cross-view feature aggregation module. Each 3D reference point constructs a set of neighboring points on its corresponding horizontal plane, where all points are projected onto the respective camera views using intrinsic parameters to extract features. These features are then aggregated through learned edge weights, effectively weighting the contributions of each neighbors. This process enables the comprehensive aggregation of 3D spatial features for each query.}
\label{fig_4}
\end{figure}

\subsection{Overall Loss}
Our training process follows the training phase of BEVFormer, sampling data from three historical timestamps within the previous two seconds to recursively generate historical BEV features $B_{t-1}$ without gradient backpropagation. At the current time $t$, the model leverages both $B_{t-1}$ and current multi-view images to capture temporal and spatial information, producing the BEV features $B_t$. These features are then passed to the detection head to predict 10 parameters for each 3D bounding box, including the center location, the scale of each box, the objects' yaw and velocity. The detector is trained end-to-end using ground-truth 3D bounding boxes. The loss used to supervise the training process includes classification loss $L_{cls}$, bounding box regression loss $L_{reg}$, and height distribution loss $L_{hgt}$:

\begin{equation}
\begin{array}{l}
L_{total}=\lambda_{1}L_{cls}+\lambda_{2}L_{reg}+\lambda_{3}L_{hgt}
\end{array}
\tag{16},
\end{equation}
where $\lambda_{1}$, $\lambda_{2}$ and $\lambda_{3}$ are the balance weights assigned to the respective loss terms. We adopt focal loss \cite{ross2017focal} and L1 loss as the loss functions for classification and regression, respectively. $L_{hgt}$ is a cross-entropy loss function defined to provide deep supervision for the estimated height distribution in the vertical adaptive height-aware sampling module (See Eq. 8). 

\begin{figure*}[!t]
\centering
\subfigure[Case 1]{
	\includegraphics[height=0.34\textheight]{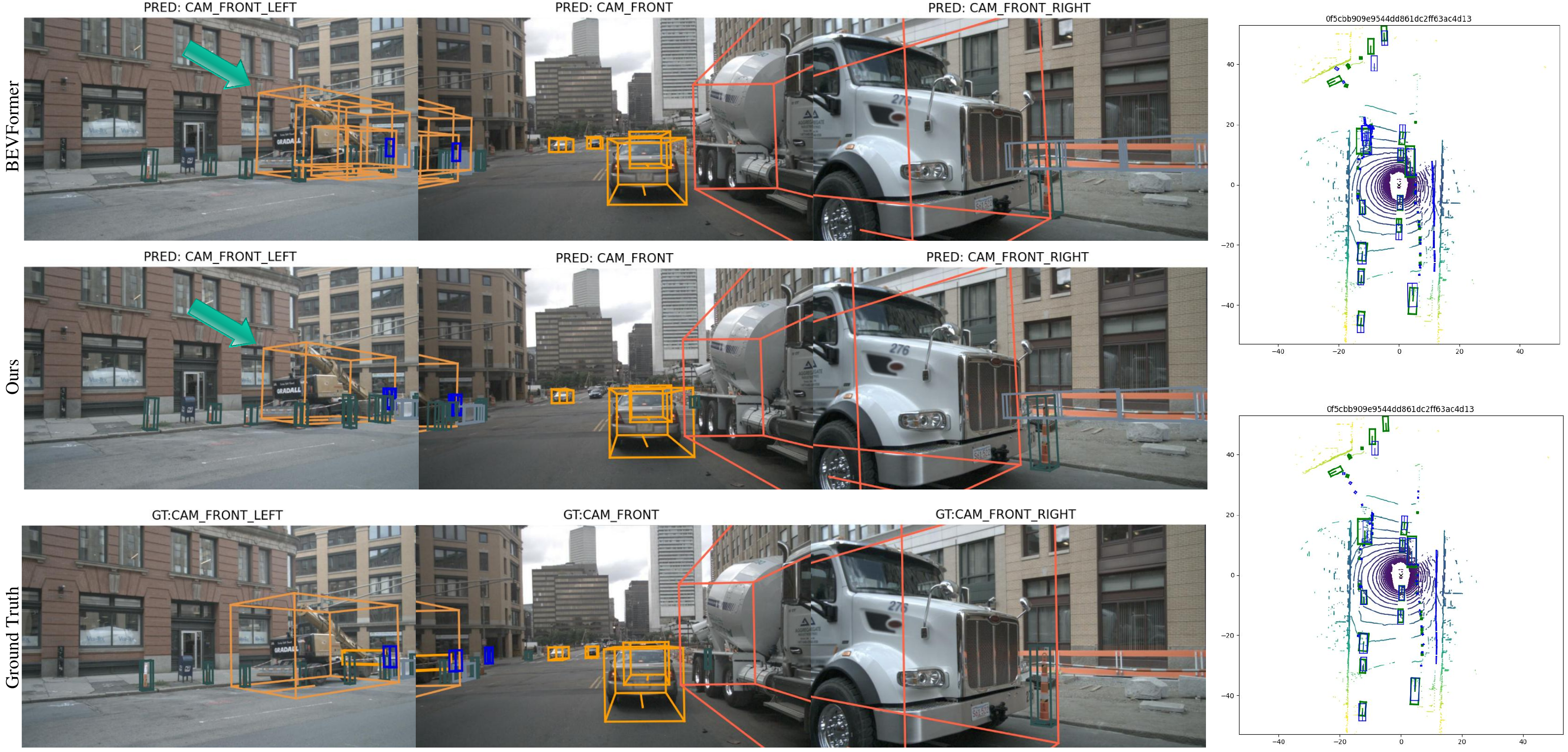}
	\label{fig:a}}
\subfigure[Case 2]{
	\includegraphics[height=0.34\textheight]{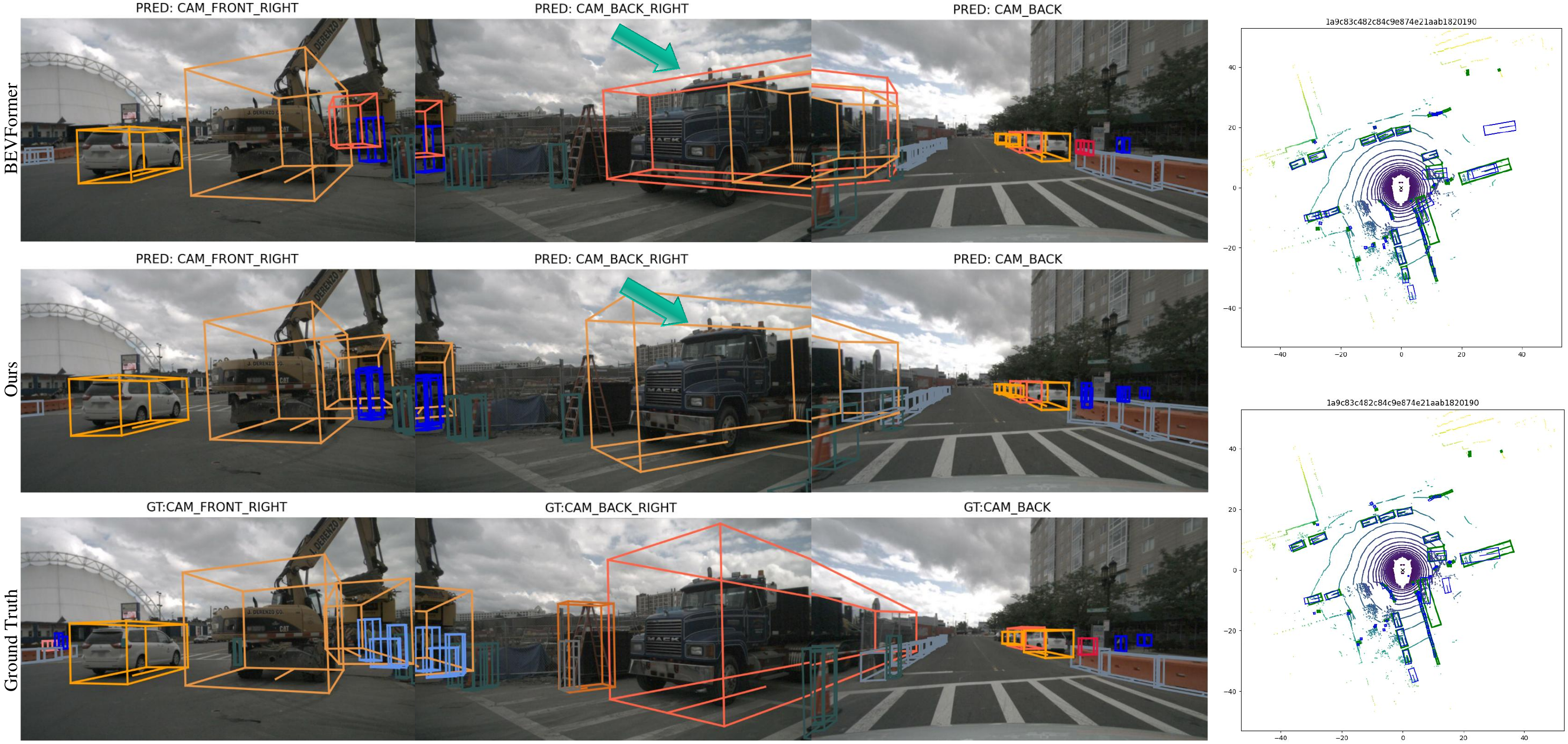}
	\label{fig:b}}
\caption{Visualization results of HV-BEV
and the baseline on nuScenes validation set. Different colors of 3D bounding boxes denote different object classes. The green arrows highlight instances where our method achieves improved detection performance on large objects. In BEVFormer, such objects are often missed or misdetected due to view truncation or their occupancy across multiple grid cells. However, our approach enhances detection by leveraging a global focus on the structured characteristics of objects, effectively mitigating these challenges. On the right are the BEV-view detection results in LiDAR point clouds, with the top showing the results of BEVFormer and the bottom showing the results of HV-BEV. Green boxes represent ground truth annotations, while blue boxes denote predicted bboxes.}
\label{fig_5}
\end{figure*}

\section{Experiments}
\subsection{Datasets and Metrics}
\textbf{The nuScenes Dataset} We train and validate our model on the publicly available nuScenes \cite{caesar2020nuscenes} autonomous driving dataset. This dataset is constructed using six cameras providing a full $360^\circ$ field of view, with images captured at a resolution of 1600$\times$900 and a frequency of 12Hz. It also incorporates LiDAR sensors operating at 20Hz with a $360^\circ$ horizontal field of view. The complete dataset consists of 1,000 diverse and challenging autonomous driving video sequences, collected from four distinct locations under varying weather and lighting conditions. Each sequence spans 20 seconds. Keyframes were sampled at a frequency of 2Hz for annotation, with 3D bounding boxes labeled for 10 object categories. Following the official dataset split, 28,130 frames from 750 sequences are used for training, while 6,019 and 6,008 frames from the remaining 300 sequences are designated for validation and testing, respectively. We evaluate the performance of our model using the 3D object detection metrics provided by the official nuScenes benchmark. These metrics include mean Average Precision (mAP), mean Average Translation Error (mATE), mean Average Scale Error (mASE), mean Average Orientation Error (mAOE), mean Average Velocity Error (mAVE), mean Average Attribute Error (mAAE), and the nuScenes Detection Score (NDS). Among these, mAP and NDS are the primary evaluation metrics.

\textbf{The Lyft Dataset} To evaluate the generalizability of the proposed method, we further trained and validated its performance on the Lyft Level 5 AV dataset \cite{kesten2019lyft}. This dataset comprises data collected using seven cameras and three LiDAR sensors, with over 55,000 manually annotated 3D bounding boxes. The data format is similar to that of the nuScenes dataset. We utilized images from six camera views—front, front-left, front-right, back, back-left, and back-right—which are consistent with those used in nuScenes, as input to our model. A total of 22,680 frames were used for training, and 27,468 frames for testing. Since the Lyft dataset does not provide an official split between training and validation sets, we followed the split scheme recommended by MMDetection3D \cite{mmdet3d2020}, using 18,900 frames for training and the remaining 3,780 frames for validation. For evaluation, we adopted a COCO-style mean Average Precision (mAP) metric, computing mAP over a range of 3D Intersection over Union (IoU) thresholds from 0.5 to 0.95. It is worth noting that an IoU threshold greater than 0.7 constitutes a stringent criterion for 3D object detection methods, which can result in relatively lower overall performance scores.

\subsection{Implementation Details}
Our model adopts BEVFormer as the baseline, with configurations tailored for three settings: $base$, $small$ and $tiny$. All settings share the same backbone, ResNet101-DCN \cite{he2016deep, dai2017deformable} and VoVnet-99\cite{lee2019energy}, initialized using pretrained parameters from FCOS3D \cite{wang2021fcos3d} and DD3D\cite{park2021pseudo}, respectively. In the $base$ setting, input images have a resolution of $1600\times900$, producing multi-scale features across four levels from the FPN, with a BEV grid size of $200\times200$, a historical sequence length of 4, and a BEV encoder comprising 6 layers. In contrast, the $small$ setting uses input images of $1280\times720$, generates single-scale feature maps, and adopts a BEV grid size of  $150\times150$, a historical sequence length of 3, and a BEV encoder with 3 layers. This design balances computational cost and performance while maintaining alignment with BEVFormer's core architecture. The $tiny$ configuration retains the same settings as $small$, except for a further reduced BEV grid size of 50×50.

For the vertical adaptive height-aware module, the number of height bins is $D=8$, and the number of 3D reference points sampled in each query is $N_{ref}=4$. The standard deviation of the Gaussian kernel used to compute the difference between the annotated box center height and the height bin values is set to $\sigma = 1$. In the DHCA module, each 3D reference point considers four neighboring nodes, and two sampling points are used for deformable attention when extracting features projected onto each 2D view. During training, the model uses the AdamW \cite{loshchilov2017decoupled} optimizer with a cosine annealing learning rate schedule \cite{loshchilov2016sgdr}. The initial learning rate is set to $2\times10^{-4}$ and the weight decay is 0.01. Training is conducted on 8 NVIDIA L20 GPUs for 24 epochs. 

\subsection{Main Results}
\textbf{Performance Improvements Over Baseline} To comprehensively evaluate the performance improvements of our model, we configured it based on the three official configuration files provided by BEVFormer (i.e., tiny, small, and base) and trained it for the same number of epochs. The performance comparison on the nuScenes validation set is shown in Tab. \uppercase\expandafter{\romannumeral1}. Our model consistently outperforms the baseline in both key metrics, mAP and NDS, highlighting the effectiveness of our proposed optimization strategies. Under the $base$ configuration, our model achieved improvements of 2.3\% in mAP and 1.6\% in NDS compared to the baseline. The FPS evaluation was consistently performed on a single NVIDIA RTX 4090 GPU. While the proposed method introduces a marginal increase in inference time over the baseline, the resulting gain in detection accuracy comes at a negligible cost to efficiency. In Fig. 5, we visualize a comparison of 3D detection results between the proposed method and the baseline on a sample of nuScenes validation set. For large objects near the ego vehicle, our method demonstrates a significant improvement in detection accuracy. The right illustrates the bird's-eye view detection results for the same sample in the LIDAR point cloud. It can be observed that our model exhibits fewer false positives and achieves a higher degree of overlap with the ground truth boxes. However, small distant objects remain challenging and are more prone to missed detections. This is primarily due to the limited visual information available for small and distant objects in the image, which leads to weak feature representations. The downsampling process in the feature extraction pipeline further degrades these representations, making it one of the key bottlenecks in vision-based 3D object detection.

\begin{table}[t]
  \centering
  \caption{Performance comparison with the baseline under three different configurations. For fairness, our model strictly follows the official configuration files provided by BEVFormer.}
  \label{tab:1}
  \setlength{\tabcolsep}{1.1mm}{
  \begin{tabular}{ccccccc}
  \toprule
  Model & Config & Grid Size & Backbone & mAP$\uparrow$ & NDS$\uparrow$ & FPS \\
  \midrule
  BEVFormer & tiny & $50\times50$ & ResNet50 & 0.252 & 0.354 & 29.75 \\
  HV-BEV(Ours) & tiny & $50\times50$ & ResNet50 & \textbf{0.284} & \textbf{0.378} & 27.91 \\
  \midrule
  BEVFormer & small & $150\times150$ & ResNet101 & 0.370 & 0.479 & 11.53 \\
  HV-BEV(Ours) & small & $150\times150$ & ResNet101 & \textbf{0.399} & \textbf{0.502} & 9.08 \\
  \midrule
  BEVFormer & base & $200\times200$ & ResNet101 & 0.416 & 0.517 & 5.83 \\
  HV-BEV(Ours) & base & $200\times200$ & ResNet101 & \textbf{0.439} & \textbf{0.533} & 4.38 \\
  \bottomrule
  \end{tabular} }
\end{table}

\textbf{Comparison with State-of-the-Arts} In addition to comparing our model with the baseline, we also present a performance comparison between our approach and some other state-of-the-art 3D odject detection methods on the nuScenes validation and testing sets. The results are summarized in Tab. \uppercase\expandafter{\romannumeral2} and Tab. \uppercase\expandafter{\romannumeral3}, respectively. On the validation set, we compared the performance of two model configurations. The $small$ configuration achieved 39.9\% mAP and 50.2\% NDS, surpassing one of the latest methods, QAF2D\cite{ji2024enhancing}, under similar settings. The $base$ configuration achieved 43.9\% mAP and 53.3\% NDS, outperforming HeightFormer, which also incorporates explicit height information, under comparable configurations. On the testing set, we evaluated our method using two mainstream backbone networks. With VoVNet-99 as the backbone, our approach achieved 50.5\% mAP and 59.8\% NDS. Notably, our model demonstrated exceptional performance on the mAAE metric, likely due to its comprehensive consideration of spatial correlations of an object. Tab. \uppercase\expandafter{\romannumeral4} reports the performance of our $base$ configuration on the validation set of the Lyft dataset. Our method consistently surpasses the baseline model and significantly outperforms several LiDAR-based 3D object detection approaches. Fig. 6 presents several robust detection results on Lyft validation set. Our model demonstrates exceptional performance in handling challenging scenarios such as large objects in cross-view and backlighting conditions on the Lyft dataset as well.

\begin{table*}[t]
  \centering
  \caption{3D object detection results on nuScenes $val$ set. *: Trained with Lidar data. \dag: Pre-trained from FCOS3D. \ddag: Pre-trained from DD3D.}
  \label{tab:2}
  \setlength{\tabcolsep}{2.8mm}{
  \begin{tabular}{c|cc|cc|ccccc}
  \toprule
  Method & Img Size & Backbone & mAP$\uparrow$ & NDS$\uparrow$ & mATE$\downarrow$ & mASE$\downarrow$ & mAOE$\downarrow$ & mAVE$\downarrow$ & mAAE$\downarrow$  \\
  \midrule
  PETR\dag\cite{liu2022petr} & $1056\times384$ & ResNet101 & 0.347 & 0.423 & 0.736 & 0.269 & 0.448 & 0.844 & 0.202 \\
  BEVFormer$_{small}$-QAF2D\dag\cite{ji2024enhancing} & $1280\times736$ & ResNet101 & 0.397 & \textbf{0.502} & \textbf{0.703} & \textbf{0.274} & \textbf{0.369} & 0.404 & 0.213 \\
  HV-BEV$_{small}$\dag(Ours) & $1280\times720$ & ResNet101 & \textbf{0.399} & \textbf{0.502} & 0.706 & 0.279 & 0.395 & \textbf{0.400} & \textbf{0.203} \\
  \midrule
  DETR3D\dag\cite{wang2022detr3d} & $1600\times900$ & ResNet101 & 0.349 & 0.434 & 0.716 & 0.268 & 0.379 & 0.842 & 0.200 \\
  BEVDet\cite{huang2021bevdet} & $1600\times640$ & Swin-T & 0.393 & 0.472 & 0.608 & \textbf{0.259} & 0.366 & 0.822 & 0.191 \\
  BEVDepth*\dag\cite{li2023bevdepth} & $1408\times512$ & ResNet101 & 0.418 & 0.538 & - & - & - & - & -  \\
  GraphDETR3D\dag\cite{chen2022graph} & $1600\times640$ & ResNet101 & 0.369 & 0.447 & - & - & - & - & - \\
  PETRv2\dag\cite{liu2023petrv2} & $1600\times640$ & ResNet101 & 0.421 & 0.524 & 0.681 & 0.267 & 0.357 & 0.377 & 0.186 \\
  OCBEV\dag\cite{qi2024ocbev} & $1600\times900$ & ResNet101 & 0.417 & 0.532 & 0.629 & 0.273 & 0.339 & 0.342 & 0.187 \\
  Sparse4D$_{T=4}$\dag\cite{lin2022sparse4d} & $1600\times900$ & ResNet101 & 0.436 & \textbf{0.541} & 0.633 & 0.279 & 0.363 & 0.317 & 0.177 \\
  HeightFormer\dag\cite{wu2024heightformer} & $1600\times900$ & ResNet101 & 0.429 & 0.532 & - & - & - & - & - \\
  BEVDet4D+WidthFormer\dag\cite{yang2024widthformer} & $1408\times512$ & ResNet101 & 0.423 & 0.531 & 0.609 & 0.269 & 0.412 & \textbf{0.302} & 0.210 \\
  Fast-BEV\dag\cite{li2024fast} & $1600\times900$ & ResNet101 & 0.413 & 0.535 & \textbf{0.584} & 0.279 & \textbf{0.311} & 0.329 & 0.206 \\
  HV-BEV$_{base}$\dag(Ours) & $1600\times900$ & ResNet101 & \textbf{0.439} & 0.533 & 0.617 & 0.264 & 0.388 & 0.375 & \textbf{0.127} \\
  \bottomrule
  \end{tabular} }
\end{table*}

\begin{table*}[t]
  \centering
  \caption{3D detection results on nuScenes $test$ set. *: Trained with Lidar data. \dag: Pre-trained from FCOS3D. \ddag: Pre-trained from DD3D.}
  \label{tab:3}
  \setlength{\tabcolsep}{3.4mm}{
  \begin{tabular}{c|cc|cc|ccccc}
  \toprule
  Method & Img Size & Backbone & mAP$\uparrow$ & NDS$\uparrow$ & mATE$\downarrow$ & mASE$\downarrow$ & mAOE$\downarrow$ & mAVE$\downarrow$ & mAAE$\downarrow$  \\
  \midrule
  BEVDet4D\cite{huang2022bevdet4d} & $1600\times640$ & Swin-T & 0.451 & \textbf{0.569} & \textbf{0.511} & \textbf{0.241} & 0.386 & \textbf{0.301} & 0.121 \\
  GraphDETR3D\dag\cite{chen2022graph} & $1600\times640$ & ResNet101 & 0.418 & 0.472 & 0.668 & 0.250 & 0.440 & 0.876 & 0.139 \\
  PETRv2\dag\cite{liu2023petrv2} & $1600\times640$ & ResNet101 & 0.456 & 0.553 & 0.601 & 0.249 & 0.391 & 0.382 & 0.123 \\
  HV-BEV$_{base}$\dag(Ours) & $1600\times900$ & ResNet101 & \textbf{0.464} & 0.556 & 0.604 & 0.261 & \textbf{0.380} & 0.393 & 0.132 \\
  \midrule
  DETR3D\ddag\cite{wang2022detr3d} & $1600\times900$ & VoVNet-99 & 0.412 & 0.479 & 0.641 & 0.255 & 0.394 & 0.845 & 0.133 \\
  BEVDepth*\ddag\cite{li2023bevdepth} & $1408\times512$ & VoVNet-99 & 0.503 & \textbf{0.600} & \textbf{0.445} & 0.245 & 0.378 & 0.320 & 0.126  \\
  GraphDETR3D\ddag\cite{chen2022graph} & $1600\times640$ & VoVNet-99 & 0.425 & 0.495 & 0.621 & 0.251 & 0.386 & 0.790 & 0.128 \\
  PETRv2\ddag\cite{liu2023petrv2} & $1600\times640$ & VoVNet-99 & 0.490 & 0.582 & 0.561 & \textbf{0.243} & 0.361 & 0.343 & 0.120 \\
  VEDet\ddag\cite{chen2023viewpoint} & $1600\times640$ & VoVNet-99 & 0.505 & 0.585 & 0.545 & 0.244 & \textbf{0.346} & 0.421 & 0.123 \\
  HeightFormer\ddag\cite{wu2024heightformer} & $1600\times900$ & VoVNet-99 & 0.481 & 0.573 & - & - & - & - & - \\
  HV-BEV$_{base}$\ddag(Ours) & $1600\times900$ & VoVNet-99 & \textbf{0.505} & 0.598 & 0.544 & 0.249 & 0.353 & \textbf{0.318} & \textbf{0.117} \\
  \bottomrule
  \end{tabular} }
\end{table*}

\begin{table}[t]
  \centering
  \caption{3D detection results on Lyft $val$ set.}
  \label{tab:4}
  \setlength{\tabcolsep}{3.5mm}{
  \begin{tabular}{cccc}
  \toprule
  Method & Modality & Backbone & mAP-3D$\uparrow$ \\
  \midrule
  Second\cite{yan2018second} & Lidar & FPN & 0.101 \\
  PointPillars\cite{lang2019pointpillars} & Lidar & FPN & 0.127 \\
  SSN\cite{zhu2020ssn} & Lidar & SECFPN & 0.121 \\
  \midrule
  BEVFormer & camera & ResNet101-FPN & 0.136 \\
  HV-BEV(Ours) & camera & ResNet101-FPN & \textbf{0.153} \\
  \bottomrule
  \end{tabular} }
\end{table}

\begin{figure}[!t]
\centering
\subfigure[Case 1]{
	\includegraphics[width=3.3in]{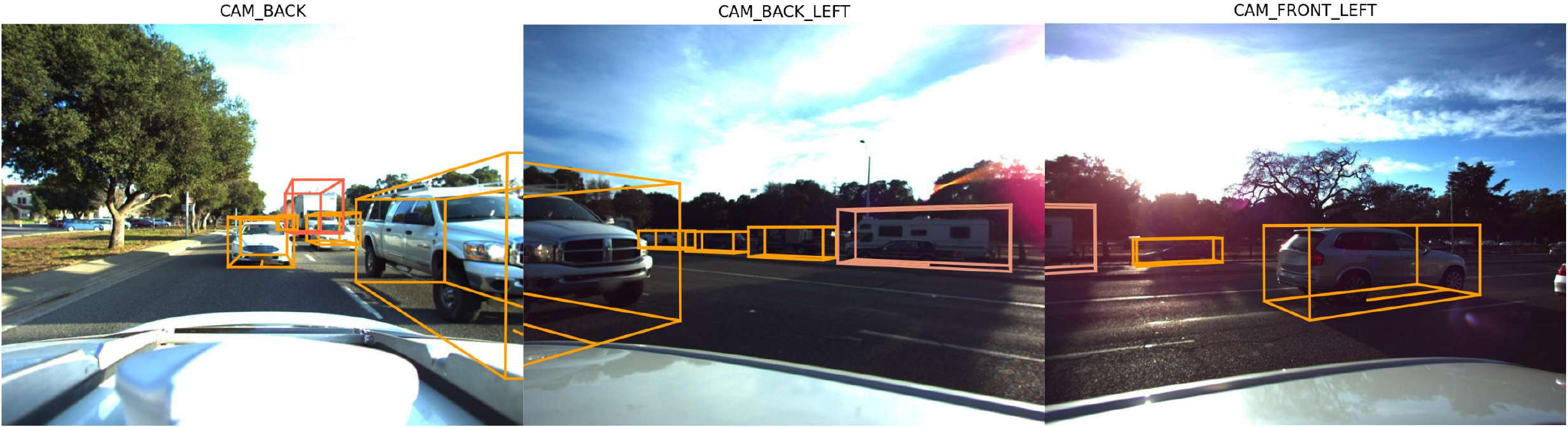}
	\label{fig:a}}
\subfigure[Case 2]{
	\includegraphics[width=3.3in]{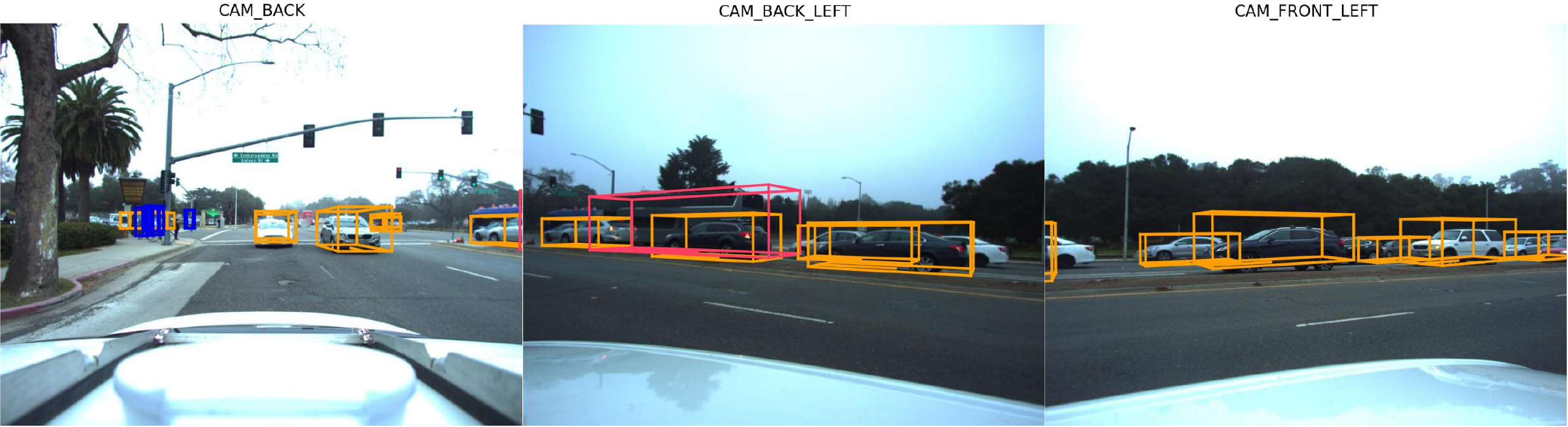}
	\label{fig:b}}
\caption{Robust visualization results on Lyft validation set. Different colors of 3D bounding boxes denote different object classes.}
\label{fig_6}
\end{figure}

\subsection{Occupancy Perception Results}
In addition to 3D object detection, we further validate the transferability of our model on the 3D occupancy prediction task. Specifically, the 256-dimensional BEV features are evenly partitioned into 16 height bins, with each bin assigned a 16-dimensional feature representation. This process transforms the BEV features into 3D volume features with an explicit height dimension. A 3D encoder-decoder network\cite{murez2020atlas} is then appended as a segmentation head to produce the final 3D semantic occupancy predictions. Moreover, we integrate our method into a BEVFormer-style 3D occupancy predictor, SurroundOcc\cite{wei2023surroundocc}. Since SurroundOcc assigns a 3D reference point to each volume query, we embed the proposed DHCA module solely into the 2D-3D cross-view attention module of SurroundOcc to enhance spatial feature aggregation. The configuration used follows the base setup described in Sec. \uppercase\expandafter{\romannumeral4}-B, where the backbone is a ResNet101-DCN pretrained on FCOS3D. Tab.  \uppercase\expandafter{\romannumeral5} reports the 3D semantic occupancy results on the nuScenes validation set. For evaluation, we adopt the IoU of occupied voxels (ignoring semantic classes) as the metric for scene completion (SC), and the mean IoU across all semantic classes as the metric for semantic scene completion (SSC). Under the simple replacement of the task head, our model outperforms BEVFormer by 1.28\% in SC IoU and 3.22\% in SSC mIoU. Additionally, incorporating the DHCA module into SurroundOcc also yields a slight improvement in overall prediction accuracy, particularly for larger objects, indicating the effectiveness of our method in enhancing feature aggregation for large-scale structures. Several qualitative comparisons are shown in Fig. 7. It can be observed that, after integrating our proposed method, the occupancy prediction model yields more complete predictions for relatively larger obstacles and exhibits improved accuracy in semantic classification. 

\begin{table*}[t]
  \centering
  \caption{3D semantic occupancy results on nuScenes $val$ set. All methods are trained with dense occupancy labels in \cite{wei2023surroundocc}.}
  \label{tab:5}
  \setlength{\tabcolsep}{1.15mm}{
  \begin{tabular}{c|cc|*{16}{c}}
  \toprule
  Model & \makecell{SC\\IoU} & \makecell{SSC\\mIoU} &
 \RotCol{barrier}{barrier} &
 \RotCol{bicycle}{bicycle} &
 \RotCol{bus}{bus} &
 \RotCol{car}{car} &
 \RotCol{const. veh.}{constveh} &
 \RotCol{motorcycle}{motorcycle} &
 \RotCol{pedestrian}{pedestrian} &
 \RotCol{traffic cone}{trafficcone} &
 \RotCol{trailer}{trailer} &
 \RotCol{truck}{truck} &
 \RotCol{drive. surf.}{drivesurf} &
 \RotCol{other flat}{otherflat} &
 \RotCol{sidewalk}{sidewalk} &
 \RotCol{terrain}{terrain} &
 \RotCol{manmade}{manmade} &
 \RotCol{vegetation}{vegetation}\\
  \midrule
  BEVFormer\cite{li2022bevformer} & 30.50 & 16.75 & 14.22 & 6.58 & 23.46 & 28.28 & 8.66 & 10.77 & 6.64 & 4.05 & 11.20 & 17.78 & 37.28 & 18.00 & 22.88 & 22.17 & 13.80 & 22.21 \\
  HV-BEV & 31.78 & 19.97 & 21.03 & 10.54 & 28.42 & 30.60 & 10.67 & 14.86 & 9.05 & 8.38 & 13.98 & 23.61 & 40.09 & 22.83 & 24.12 & 24.30 & 14.69 & 22.27 \\
  SurroundOcc\cite{wei2023surroundocc} & 31.49 & 20.30 & 20.59 & \textbf{11.68} & 28.06 & 30.86 & \textbf{10.70} & 15.14 & \textbf{14.09} & \textbf{12.06} & \textbf{14.38} & 22.06 & 37.29 & \textbf{23.70} & 24.49 & 22.77 & 14.89 & 21.86 \\
  SurroundOcc-DHCA & \textbf{32.93} & \textbf{20.97} & \textbf{22.14} & 11.02 & \textbf{29.57} & \textbf{31.82} & 10.68 & \textbf{15.26} & 13.30 & 11.30 & 13.89 & \textbf{23.81} & \textbf{40.40} & 23.19 & \textbf{26.37} & \textbf{24.69} & \textbf{15.30} & \textbf{22.76} \\
  \bottomrule
  \end{tabular} }
\end{table*}

\begin{figure*}[!t]
\centering
\subfigure[Case 1]{
	\includegraphics[height=0.245\textheight]{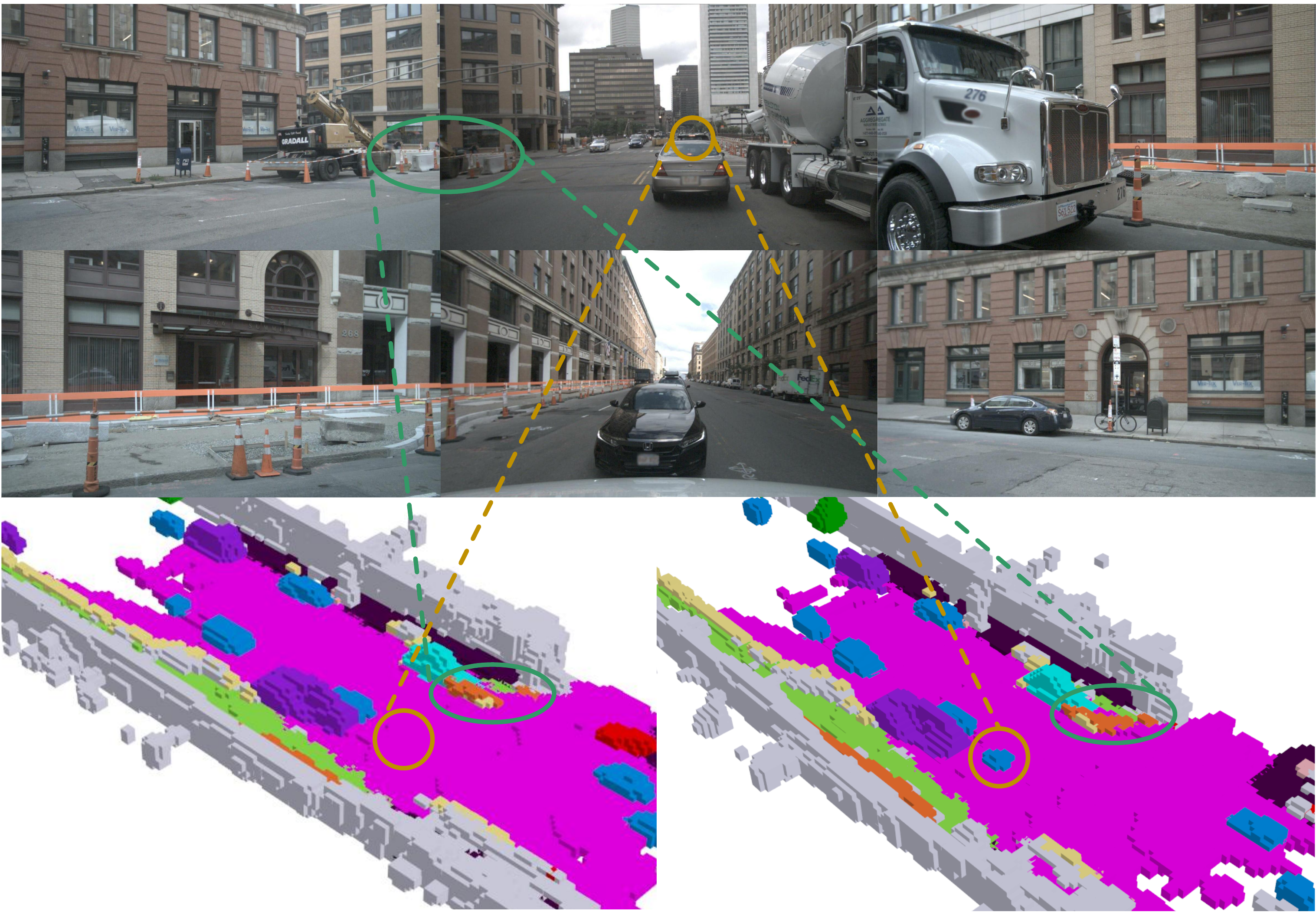}
	\label{fig:a}}
\subfigure[Case 2]{
	\includegraphics[height=0.245\textheight]{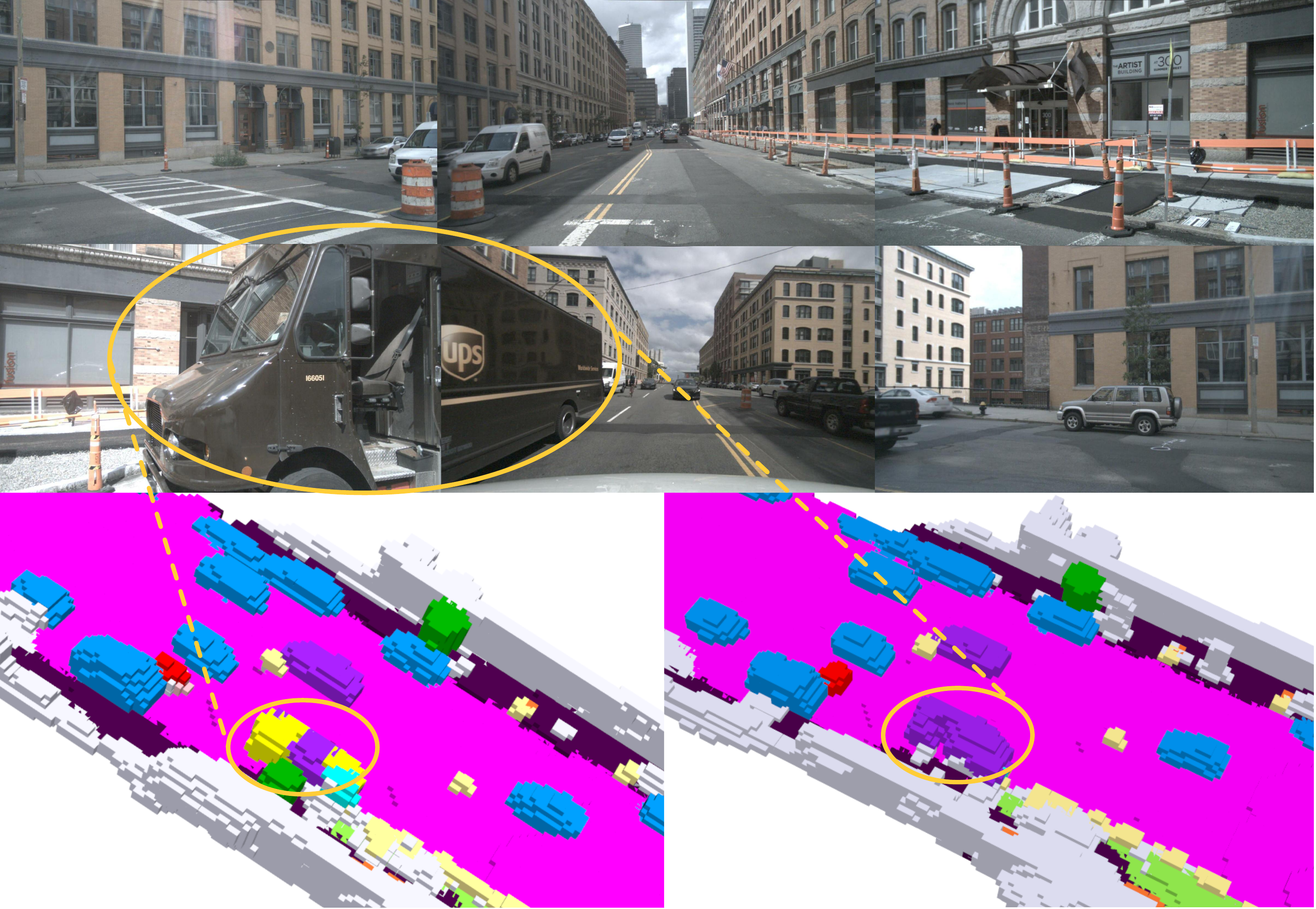}
	\label{fig:b}}
\caption{A qualitative comparison of 3D semantic occupancy prediction results between SurroundOcc and SurroundOcc-DHCA on two samples from the nuScenes validation set. The bottom-left corner shows the predictions from SurroundOcc, while the bottom-right corner presents the results from SurroundOcc-DHCA. Colored elliptical boxes highlight regions where the prediction performance of the two methods differs significantly.}
\label{fig_7}
\end{figure*}

\subsection{Ablation Study}
In order to gain deeper insights into the effectiveness of the proposed method's individual components, we conducted a series of ablation studies on the nuScenes validation set using the $small$ configuration of our model. 

\textbf{Main Ablations} In this subsection, we focus on validating the effectiveness of the proposed enhancement strategies. As shown in Tab. \uppercase\expandafter{\romannumeral6}, the vanilla BEVFormer achieves an mAP of 37.0\% and an NDS of 47.9\%. Incorporating the Vertical Adaptive Height-Aware (VHA) sampling strategy improves these metrics by 0.5\% and 0.3\%, respectively. Adding the Dynamic Horizontal Cross-view Feature Aggregation (DHCA) module further boosts mAP and NDS by 2.2\% and 1.9\%, respectively. When both strategies are applied simultaneously, the mAP and NDS reach 39.9\% and 50.2\%, representing gains of 2.9\% and 2.3\%, respectively. Additionally, significant improvements are observed across other metrics as well.

\begin{table*}[t]
  \centering
  \caption{Main ablation results of HV-BEV on nuScenes val set. "VHA" and "DHCA" denotes for vertical adaptive height-aware sampling and dynamic horizontal cross-view feature aggregation, respectively.}
  \label{tab:6}
  \setlength{\tabcolsep}{6mm}{
  \begin{tabular}{cc|cc|ccccc}
  \toprule
  VHA & DHCA & mAP$\uparrow$ & NDS$\uparrow$ & mATE$\downarrow$ & mASE$\downarrow$ & mAOE$\downarrow$ & mAVE$\downarrow$ & mAAE$\downarrow$  \\
  \midrule
  \resizebox{!}{\fontcharht\font`M}{\input{usym2717.tikz}} & \resizebox{!}{\fontcharht\font`M}{\input{usym2717.tikz}} & 0.370 & 0.479 & 0.721 & 0.279 & 0.407 & 0.436 & 0.220 \\
  \resizebox{!}{\fontcharht\font`M}{\input{usym2713.tikz}} & \resizebox{!}{\fontcharht\font`M}{\input{usym2717.tikz}} & 0.375 & 0.482 & 0.716 & 0.280 & 0.405 & 0.431 & 0.217 \\
  \resizebox{!}{\fontcharht\font`M}{\input{usym2717.tikz}} & \resizebox{!}{\fontcharht\font`M}{\input{usym2713.tikz}} & 0.392 & 0.498 & 0.709 & \textbf{0.279} & 0.399 & 0.409 & 0.208 \\
  \resizebox{!}{\fontcharht\font`M}{\input{usym2713.tikz}} & \resizebox{!}{\fontcharht\font`M}{\input{usym2713.tikz}} & \textbf{0.399} & \textbf{0.502} & \textbf{0.706} & \textbf{0.279} & \textbf{0.395} & 0.\textbf{400} & \textbf{0.203} \\
  \bottomrule
  \end{tabular} }
\end{table*}

\textbf{Ablations on VHA} In the VHA module, the number of height bins determines the sampling density of 3D reference points. Setting too few height bins (e.g., 4) degrade the model into uniform height sampling, similar to the vanilla BEVFormer. Conversely, an excessive number of height bins may cause reference points to overly concentrate within a narrow height range. Therefore, selecting an optimal number of height bins is crucial. Tab. \uppercase\expandafter{\romannumeral7} summarizes the experimental results for different height bin configurations. When the number of height bins is set to 8, the model achieves the best performance. The performance with 6 bins is only slightly lower than with 8 bins, but increasing the number of bins beyond 8 leads to a notable decline in performance. We infer that this decline occurs because the supervision for the height distribution prediction is based on the 3D center of the object. As the number of bins increases and becomes denser, the 3D reference points are sampled closer to the object center, resulting in higher probabilities being assigned to bins near the center. This concentration on the central region may force the model to overlook the broader height range, thereby degrading overall performance. Although this issue can be alleviated by increasing the number of $N_{ref}$, doing so inevitably incurs higher computational overhead, as each 3D reference point and its neighboring points require feature sampling from the image. Moreover, overly dense sampling may lead to performance degradation due to over-smoothing.

\begin{table}[t]
  \centering
  \caption{Effectiveness of the number of height bins in VHA module on nuScenes val set.}
  \label{tab:7}
  \setlength{\tabcolsep}{2.5mm}{
  \begin{tabular}{ccccc}
  \toprule
  Height Bin Num & mAP$\uparrow$ & NDS$\uparrow$ & mATE$\downarrow$ & mAOE$\downarrow$ \\
  \midrule
  4 & 0.392 & 0.498 & 0.709 & 0.398 \\
  6 & 0.397 & 0.501 & \textbf{0.706} & 0.396 \\
  8 & \textbf{0.399} & \textbf{0.502} & \textbf{0.706} & \textbf{0.395} \\
  10 & 0.381 & 0.486 & 0.719 & 0.411 \\
  12 & 0.372 & 0.476 & 0.760 & 0.419 \\
  \bottomrule
  \end{tabular} }
\end{table}

\textbf{Ablations on DHCA} In the DHCA module, each 3D reference point constructs a horizontal dynamic graph centered on itself, with each node in the graph aggregating relevant features from its respective location. Theoretically, increasing the number of neighboring nodes allows the aggregation of richer features but also incurs higher computational costs. Determining the optimal number of neighboring nodes to balance accuracy and computational efficiency is a critical consideration. Tab. \uppercase\expandafter{\romannumeral8} presents a comparison of the model's accuracy, memory consumption, and inference speed under different settings of the number of neighboring nodes. Interestingly, performance degrades when the number of neighbors becomes excessively large, potentially due to over-smoothing. The inference speed remains comparable when the number of neighbors is set to 2 or 4; however, a significant drop in FPS is observed when the number exceeds 8. We adopt 4 neighbors as the default configuration, which not only reduces memory usage but also maintains competitive accuracy.

\begin{table}[t]
  \centering
  \caption{Effectiveness of the number of neighbors in DHCA module on nuScenes val set.}
  \label{tab:8}
  \setlength{\tabcolsep}{3.8mm}{
  \begin{tabular}{ccccc}
  \toprule
  Neighbor Num & mAP$\uparrow$ & NDS$\uparrow$ & Memory & FPS \\
  \midrule
  2 & 0.386 & 0.489 & $\sim11700$M & 9.87 \\
  4 & 0.399 & 0.502 & $\sim13900$M & 9.08 \\
  8 & 0.402 & 0.504 & $\sim17150$M & 7.25 \\
  16 & 0.379 & 0.480 & $\sim22950$M & 3.04 \\
  \bottomrule
  \end{tabular} }
\end{table}

\textbf{Effectiveness of decoupled sampling method} This study enhances feature aggregation through height-aware reference point sampling along the vertical axis and graph-based cross-view perception in the horizontal plane. However, it is generally acknowledged that increasing the number of reference points for a single query, without considering computational cost, inherently allows the sampling of more image features, thereby improving detection accuracy. The purpose of this section is to validate the effectiveness of the proposed feature sampling strategies, rather than attributing performance improvements solely to the increased number of reference points. Tab. \uppercase\expandafter{\romannumeral9} reports the results of experiments conducted on the baseline model under the $small$ configuration. These tests examine the effect of increasing the number of uniformly sampled 3D reference points per query ($N_{ref} = 4, 8, 16$) on detection accuracy. The results indicate that increasing $N_{ref}$ to 8 yields only minor performance improvements, while further increasing it to 16 results in a decline across all metrics. We attribute this decline to the potential redundancy or irrelevance introduced by simply adding more uniformly sampled reference points, which may weaken the impact of critical features and disrupt the model’s learning process. These findings further validate the effectiveness of the proposed decoupled horizontal and vertical feature sampling strategies.

\begin{table}[t]
  \centering
  \caption{Effectiveness of increasing the number of 3D reference points within each query on nuScenes val set.}
  \label{tab:9}
  \setlength{\tabcolsep}{3.5mm}{
  \begin{tabular}{ll|c|cc}
  \toprule
  Model & Config & $N_{ref}$ & mAP$\uparrow$ & NDS$\uparrow$ \\
  \midrule
  \multirow{3}{3.5mm}{BEVFormer} & \multirow{3}{3.5mm}{\centering small} &
       4 & 0.370 & 0.479 \\
   & & 8 & 0.374 & 0.481 \\
   & & 16 & 0.351 & 0.462 \\
   \midrule
   HV-BEV(Ours) & small & 4 ($\times4$ nbs) & \textbf{0.399} & \textbf{0.502} \\
  \bottomrule
  \end{tabular} }
\end{table}

\section{Conclusion}
In this paper, we propose HV-BEV, a novel dense BEV-based 3D object detection framework that decouples horizontal and vertical feature sampling. The framework leverages adaptive height-aware 3D reference point sampling by integrating historical and current BEV height distributions to focus on object-relevant height ranges. It then learns a graph structure on the horizontal plane at the reference points’ corresponding heights, enabling the aggregation of structured spatial information for objects in 3D space. Finally, we introduce a loss function with depth supervision, utilizing the ground-truth center height of objects to refine the predicted height distribution. Extensive experiments on the nuScenes dataset validate the effectiveness of the proposed approach. 

Our approach aims to serve as a reference for dense BEV-based 3D perception tasks beyond object detection. However, while our method significantly improves detection accuracy, it shares a limitation with the baseline model (i.e., BEVFormer): high computational cost. In future work, we plan to explore hybrid perception models that combine dense and sparse BEV representations to further optimize model performance while reducing computational overhead.

\vfill

\end{document}